\documentclass[sigconf,nonacm,screen]{acmart}

\usepackage{longtable}
\usepackage{array}
\usepackage{calc}
\usepackage{tikz}
\usetikzlibrary{positioning, arrows.meta}

\title{A Query Engine for the Agents}

\author{Kenny Daniel}
\affiliation{%
  \institution{Hyperparam}
  \city{Seattle}
  \country{USA}
}
\email{kenny@hyperparam.app}

\begin{document}

\begin{abstract}
The fastest-growing data in production today is unstructured text: agent
traces, chat logs, reasoning chains, model outputs. People want to
analyze it, and the questions worth asking (``show me where the agent
got confused'') cannot be answered by SQL alone, since text is not
queryable without a model in the query path. The natural place this
analysis is happening is the new class of AI applications, Claude Code,
Cursor, Claude Desktop, and in-browser agents, that run client-side and
host both a human user and an LLM agent in the same process. These
applications increasingly want to work with data, but the lakehouse read
path has been hard to use from a JS runtime: Spark, Trino, and managed
warehouses do not fit there. To build this new kind of AI data
application, three properties of the engine become first-order: a
JS-native distribution that drops into the runtime the application
already runs in, a bundle small enough to ship inside a cold tab or
per-turn agent sandbox, and a way to interleave analytic operators with
model-based interpretation of text. We present \textbf{Hyperparam},
three open-source JavaScript libraries (Hyparquet, Squirreling, Icebird)
totaling under 70 KB, that read Parquet and Apache Iceberg directly from
object storage and meet the third property with per-cell, async-native
SQL execution, so expensive cells fire only when downstream operators
demand them. Squirreling runs LLM-shaped async UDFs over 300× faster
than DuckDB-WASM on filter-bounded queries (and 192× on sort-bounded
queries) and completes a ten-task agent analyst suite at two-thirds
lower cost. We argue that data engineering as a discipline needs to
update for the AI-native client applications now in production and the
agents that work alongside their users.


\end{abstract}

\maketitle

\hypertarget{introduction}{%
\section{Introduction}\label{introduction}}

The data that practitioners and agents most want to query is
unstructured text, written by software: chat logs
\citep{he2021automatedlog}, tool-use traces \citep{dong2024agentops},
reasoning chains, coding-agent logs, and model outputs, at millions to
billions of records per tenant \citep{deepview2026}.
\citet{liu2025agentic} project that AI agents will become the dominant
consumer of this data, issuing orders of magnitude more probes than
humans; they report that, at Neon, agents created 20× more database
branches and performed 50× more rollbacks relative to humans.

Classic OLAP was not built for this shape of data. The questions that
surface real problems, like where users got frustrated, why a tool call
failed, or where an agent went down a rabbit hole, cannot be answered by
SQL alone \citep{john2023datachat}; they require a model in the query to
interpret each row, with \texttt{llm()} UDFs as first-class operators
alongside scans, joins, and aggregates, and cheaply, since each call is
a billable inference. The data is heavy and its owners want it to stay
where it already lives \citep{armbrust2021lakehouse}. The conventional
read path puts heavy infrastructure (Spark, Trino, a managed warehouse,
a query gateway) between consumer and data, slowing the
curate-review-act loop \citep{alspaugh2014loganalysis}.

The natural place this analysis is happening is a new class of AI
applications that has reached production. Claude Code (Bun CLI), Claude
Desktop, Cursor, and VS Code (host to Copilot, Cline, and Continue) ship
as Electron; hyperparam.app runs in a browser tab; per-turn agent
sandboxes are Node \texttt{vm} contexts, Deno, or Web Workers. They
share three properties: client-side, often JavaScript, and a process
that hosts both a human user and an LLM agent. They increasingly want to
query data, but the lakehouse read path does not fit a JS runtime:
Spark, Trino, and managed warehouses are server-side. An agent issuing
exploratory probes inside a tool-use turn cannot wait on a credentialed
backend on the critical path either. The engine has to come to the
consumer.

\textbf{Our position.} Three properties of the engine then become
first-order: a JS-native distribution, so the engine drops into the
runtime the application already runs in with no second runtime or FFI
between user and agent; a bundle small enough to load inside a cold tab
or a per-turn agent sandbox; and a way to interleave analytic operators
with model-based interpretation of text, since trace cells can be tens
of kilobytes of compressed text and an \texttt{llm()} UDF call is a
multi-second remote inference with real dollar cost. Hyperparam meets
the third with per-cell, async-native execution, where expensive cells
fire only when downstream operators demand them and the query path can
\texttt{await} model APIs as a first-class step. We ground the position
in three open-source JavaScript libraries (Hyparquet, Squirreling,
Icebird) totaling under 70 KB; DuckDB-WASM \citep{kohn2022duckdbwasm} is
the closest prior art and we contrast in §2.

\hypertarget{related-work}{%
\section{Related Work}\label{related-work}}

\textbf{Lakehouse foundations.} The lakehouse architecture
\citep{armbrust2021lakehouse} places a transactional metadata layer such
as Apache Iceberg, Delta Lake, or Hudi on object storage, typically over
Parquet \citep{vohra2016parquet}. Hyperparam reads both layers directly;
the storage substrate is present in every production deployment we
target.

\textbf{In-place and embedded analytics.} DuckDB
\citep{raasveldt2019duckdb} is an embeddable OLAP engine designed to run
inside a host process; DuckDB-WASM \citep{kohn2022duckdbwasm} compiles
it to WebAssembly and was ahead of the curve in establishing that
interactive analytics belong in the browser. It is the prior art closest
in spirit to Hyperparam, and we extend the thesis into the
unstructured-text and agent regime on two axes. First, bundle size.
DuckDB-WASM's core WebAssembly module is roughly 8 MB gzipped over the
wire (35 MB uncompressed), plus a worker shim and JS glue; Hyparquet,
Squirreling, and Icebird together ship in under 70 KB gzipped, a roughly
100× gap on the same axis that matters for cold-tab page loads from
incident links and for per-turn spin-up inside an agent sandbox. Second,
the unit of laziness. In the DuckDB-WASM path we test, execution is
vectorized in morsels, but scalar UDF calls still cross the WASM
boundary synchronously from the engine's perspective: a long-running
awaited call inside a UDF such as an \texttt{llm()} inference stalls
that path rather than letting other rows continue around it.
Squirreling's AsyncGenerator operators stream \emph{deferred cells}
throughout, so any cell can be an awaited remote or inference call, and
cells that no downstream operator reads are never decompressed (§3.2).
DuckDB-WASM remains the right tool for many workloads; the agentic-text
loop is where the gap is widest.

Other JavaScript Parquet tooling does not address this workload.
Arrow-JS \citep{apache_arrow} is oriented around Arrow IPC buffers and
does not do HTTP-range reads over remote Parquet, \texttt{parquetjs} is
unmaintained, DataFusion lacks a production-ready WASM package, and
Arquero \citep{moritz2020arquero} operates over in-memory DataFrames and
complements a storage access layer rather than replacing one.

\textbf{Agentic data systems.} \citet{liu2025agentic} argue that the
shift from human operators to AI agents fundamentally breaks existing
data system assumptions, since agents issue orders of magnitude more
probes, most redundant and speculative, and the query layer must become
satisficing-aware. \citet{tagliabue2025lakehouse} argue from the write
side, where the agentic lakehouse needs transaction isolation and
governance primitives for concurrent agent writes.
\citet{zaharia2024compound} motivate the broader shift from monolithic
models to compound AI systems. A parallel thread treats LLM-based
operators as first-class query primitives, with accuracy guarantees
\citep{patel2025semanticoptimization}, cost-aware optimization over
unstructured collections \citep{liu2025palimpzest}, and dedicated
engines for semantic predicates \citep{dai2024uqe}. Hyperparam addresses
the read seam these framings leave open. How does a user, or an agent,
get to the data once it has been stored?

\textbf{Trace observability platforms.} Langfuse, Braintrust, LangSmith,
Helicone, and Phoenix address the same user-facing workload by holding
traces in a store they operate, on top of which they build curation and
evaluation UIs. Hyperparam's stance is different: trace data stays with
the owner in their existing storage, and the same JavaScript engine runs
in whatever runtime the consumer is already in (browser tab, agent
sandbox, or one tab hosting both), so the iteration loop closes inside
one engine rather than being split between a vendor UI for the human and
a hosted API for the agent.

\hypertarget{the-hyperparam-stack}{%
\section{The Hyperparam Stack}\label{the-hyperparam-stack}}

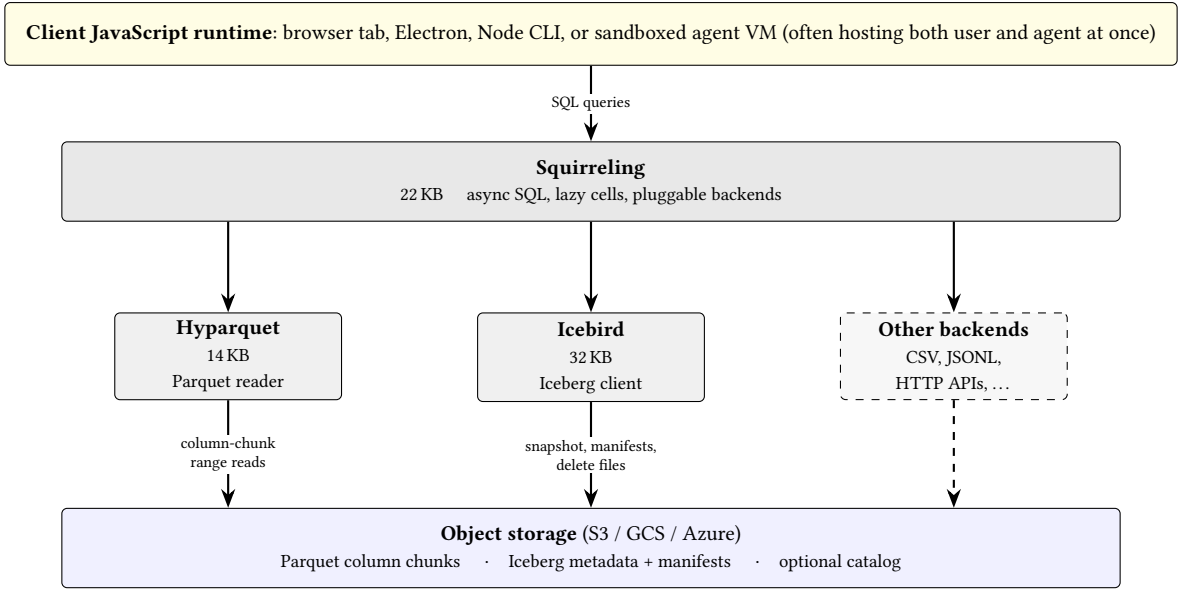
\begin{figure*}[t]
\centering
\begin{tikzpicture}[
  font=\small,
  host/.style={draw, rounded corners=2pt, fill=yellow!12, align=center, inner sep=8pt, minimum width=14cm},
  engine/.style={draw, rounded corners=2pt, fill=gray!18, align=center, minimum width=14cm, minimum height=1.05cm},
  lib/.style={draw, rounded corners=2pt, fill=gray!12, align=center, minimum width=3cm, minimum height=1.05cm},
  other/.style={draw, dashed, rounded corners=2pt, fill=gray!6, align=center, minimum width=3cm, minimum height=1.05cm},
  store/.style={draw, rounded corners=2pt, fill=blue!6, align=center, minimum width=14cm, minimum height=1.05cm},
  arr/.style={-{Stealth[length=2.5mm]}, thick},
  ann/.style={font=\scriptsize, align=center, fill=white, inner sep=1pt},
]
\node[host] (host) {\textbf{Client JavaScript runtime}: browser tab, Electron, Node CLI, or sandboxed agent VM (often hosting both user and agent at once)};

\node[engine, below=1.0cm of host.south] (sq) {\textbf{Squirreling}\\\footnotesize 22\,KB \quad async SQL, lazy cells, pluggable backends};

\draw[arr] (host.south) -- (sq.north) node[ann, midway] {SQL queries};

\node[lib, below=1.2cm of sq.south, xshift=-4.8cm] (hp) {\textbf{Hyparquet}\\\footnotesize 14\,KB\\\footnotesize Parquet reader};
\node[lib, below=1.2cm of sq.south]                (ic) {\textbf{Icebird}\\\footnotesize 32\,KB\\\footnotesize Iceberg client};
\node[other, below=1.2cm of sq.south, xshift=4.8cm] (ot) {\textbf{Other backends}\\\footnotesize CSV, JSONL,\\\footnotesize HTTP APIs, \ldots};

\draw[arr] (sq.south -| hp) -- (hp.north);
\draw[arr] (sq.south -| ic) -- (ic.north);
\draw[arr] (sq.south -| ot) -- (ot.north);

\node[store, below=1.4cm of ic] (os) {\textbf{Object storage} (S3 / GCS / Azure)\\\footnotesize Parquet column chunks \quad$\cdot$\quad Iceberg metadata + manifests \quad$\cdot$\quad optional catalog};

\draw[arr] (hp.south) -- (hp.south |- os.north) node[ann, midway] {column-chunk\\range reads};
\draw[arr] (ic.south) -- (ic.south |- os.north) node[ann, midway] {snapshot, manifests,\\delete files};
\draw[arr, dashed] (ot.south) -- (ot.south |- os.north);
\end{tikzpicture}
\caption{The Hyperparam stack: a client JS runtime issues SQL to Squirreling, which dispatches reads through pluggable backends (Hyparquet for Parquet, Icebird for Iceberg, others for CSV / JSONL / HTTP APIs) that fetch directly from object storage via a custom-fetch hook supplying credentials. Under 70\,KB combined, gzipped. No backend service mediates between consumer and data.}
\label{fig:stack}
\end{figure*}

Hyperparam is a browser-native lakehouse client built around
Squirreling, a small async SQL engine with pluggable backends. The
client issues SQL to Squirreling, which dispatches scans to whichever
backend matches the source: Hyparquet for Parquet files, Icebird for
Iceberg tables. Row-oriented adapters are available for CSV, JSONL, or
HTTP APIs, but we focus on the column-oriented backends, as they offer
the most performance benefits, and are widely used for large-scale
storage. Each library is independently deployable, and together they
turn a URL plus object-storage credentials into a live query target.
Figure\textasciitilde{}\ref{fig:stack} shows the composition.

\hypertarget{hyparquet-range-efficient-parquet-reads}{%
\subsection{Hyparquet: Range-Efficient Parquet
Reads}\label{hyparquet-range-efficient-parquet-reads}}

Hyparquet reads the Parquet footer via a speculative trailing range
request (a HEAD-derived absolute byte offset, since browser CORS blocks
the suffix-range form \texttt{Range:\ bytes=-N}), decodes the metadata,
and issues parallel range requests for exactly the column chunks needed
by the query, skipping row groups via min/max statistics. No full-file
download occurs. At 14 KB gzipped with zero dependencies, it implements
the full Parquet specification including nested types via Dremel-encoded
repetition and definition levels \citep{melnik2010dremel}, which matters
for deeply nested LLM trace schemas. The practical effect on the
consumer is fast spin-up: an incident link can open a 40 GB Iceberg
table of agent traces on S3 and return the first row in under a second
from a cold tab, with no \texttt{pip\ install}, cluster, or query
gateway in the path. The cold-start gap is not our main claim, since
most enterprise Trino and Spark deployments are pre-provisioned; the
more durable comparison is warm-start, where each cluster probe still
incurs a coordinator round trip and worker-side scan that the in-browser
path avoids (Table 1).

\begin{table}[h]\centering
\begin{tabular}{@{}lll@{}}
\toprule
Stack & Cold & Warm \\
\midrule
Hyperparam & 0.6 s & 0.2 s \\
DuckDB-WASM & 19 s & 1.3 s \\
Trino on EC2 & \textasciitilde3 min & 1.1 s \\
Spark on EMR & \textasciitilde8 min & 3.4 s \\
\bottomrule
\end{tabular}\end{table}

\emph{Table 1: time-to-first-row for}
\texttt{SELECT\ *\ FROM\ traces\ WHERE\ session\_id\ =\ ?\ LIMIT\ 50}
\emph{on a 40 GB Iceberg table of agent traces on S3. ``Cold'' includes
bundle fetch and engine init for browser stacks and cluster provision
for Trino/Spark; ``warm'' assumes the engine is already up and
footer/manifests are cached. DuckDB-WASM 1.33.1-dev45.0. Means across
repeated trials.}

\hypertarget{squirreling-async-native-sql-with-per-cell-laziness}{%
\subsection{Squirreling: Async-Native SQL with Per-Cell
Laziness}\label{squirreling-async-native-sql-with-per-cell-laziness}}

Squirreling is a from-scratch JavaScript SQL engine, 22 KB minified and
gzipped, with zero dependencies, whose execution primitive is a
\emph{lazy deferred cell}. A cell encapsulates a computation such as a
column decompression, a network fetch, or an LLM inference call, and is
not evaluated until its value is demanded by a downstream operator or a
\texttt{LIMIT} boundary. Its SQL dialect is deliberately permissive: the
target consumer is a model rather than a fixed API contract, so the
parser accepts aliases and idioms from multiple SQL dialects. Joins can
span heterogeneous backends, e.g.~an Iceberg table against a JSONL file
in one query.

Squirreling's cell-level pull aligns cost with demand: a column chunk is
not fetched until some row needs it, a value is decompressed only for
rows that pass the predicate and reach a projecting operator, and an
\texttt{llm()} call is not issued until the row reaches the output. This
extends the late-materialization insight of
\citet{abadi2007materialization} beyond the column-store bandwidth
argument to network I/O and LLM cost.

A query
\texttt{SELECT\ llm(\textquotesingle{}classify\textquotesingle{},\ content)\ FROM\ traces\ WHERE\ session\_id\ =\ \$id\ LIMIT\ 5}
therefore makes exactly five inference calls, regardless of how many
rows match the predicate before the limit. This guarantee holds on the
streaming path taken by queries without ORDER BY, GROUP BY, or
aggregates; queries that require a global sort or group switch to a
buffered path, at the cost of the per-cell bound. On the streaming path,
satisficing \citep{liu2025agentic} becomes a query-time primitive. The
consumer stops when enough rows have been seen, and the system charges
only for what was read.

The async-native execution boundary is the other key property.
Squirreling's operators are composed as JavaScript AsyncGenerators
throughout. Every operator yields rows as they become available, every
blocking operation is an awaited Promise, and the planner reasons over
the dependency graph of deferred cells without a thread-pool
abstraction. The \texttt{llm()} UDF accepts a model identifier, a prompt
template, and column references; results stream back as they arrive,
partial results are visible before completion, and the consumer can
cancel mid-query. Human judgment stays in the loop, since
\citet{shankar2024validators} show that LLM judges diverge
systematically from human preference, so the UDF is a filter that
surfaces rows worth reviewing, not a replacement.

We measure the practical effect in a head-to-head with DuckDB-WASM, the
closest browser-native baseline (Table 2). The benchmark uses a mock
\texttt{llm()} UDF with a 5 ms per-call delay, which is much faster than
real hosted LLM inference; the point is to isolate call counts and async
pipelining without making the experiment slow to run. Two independent
properties drive the gap. \emph{Plan-shape laziness} governs call count:
on shape B, Squirreling's cell-level streaming short-circuits at the
LIMIT (606 calls at N=50, 12.5\% selectivity), while DuckDB-WASM cannot
push LIMIT past a filter on a generated column and evaluates in
2,048-row morsels. On shape C, both engines must evaluate every row
before the sort bounds the output, and call counts converge. \emph{Async
pipelining} governs wall-clock independent of call count: Squirreling's
AsyncGenerator operators overlap up to 256 concurrent invocations, while
DuckDB-WASM's \texttt{createScalarFunction} is a synchronous C++ call
into WASM and pins concurrency at one. Registering an \texttt{async}
callback does not recover it; DuckDB-WASM reads the unresolved Promise
as the declared Arrow type and writes a zero into every result cell, so
async UDFs return silently wrong answers rather than slow correct ones.
On shape C, call-count parity still leaves a 192× wall-clock gap purely
from pipelining.

\begin{table}[h]\centering
\begin{tabular}{@{}lll@{}}
\toprule
Shape & DuckDB-WASM & Squirreling \\
\midrule
A: \texttt{LIMIT N}                   & \shortstack[l]{50 calls\\320 ms}     & \shortstack[l]{50 calls\\20 ms}    \\
\midrule
B: \texttt{WHERE llm(...)=1 LIMIT N}  & \shortstack[l]{2,048 calls\\12,620 ms} & \shortstack[l]{606 calls\\40 ms}   \\
\midrule
C: \texttt{ORDER BY llm(...) LIMIT N} & \shortstack[l]{10,050 calls\\61,667 ms} & \shortstack[l]{10,000 calls\\321 ms} \\
\midrule
D: \texttt{WHERE native LIMIT N}      & \shortstack[l]{50 calls\\333 ms}     & \shortstack[l]{50 calls\\24 ms}    \\
\bottomrule
\end{tabular}
\end{table}

\emph{Table 2: UDF call count and wall-clock at N=50 on a 10,000-row
input with a 5 ms per-call mock latency, across four query shapes.
Squirreling 0.12.16; DuckDB-WASM 1.33.1-dev45.0. Call counts are
deterministic given the input; wall-clock is the median across 5
trials.}

\hypertarget{icebird-client-side-iceberg}{%
\subsection{Icebird: Client-Side
Iceberg}\label{icebird-client-side-iceberg}}

Icebird \citep{iceberg_spec} resolves Iceberg snapshots client-side into
the set of Parquet files Hyparquet then reads, applying position and
equality deletes in the client. The central design decision is the
custom-fetch hook. Callers inject a \texttt{fetch}-compatible function
that supplies credentials such as AWS SigV4, GCS HMAC, or bearer tokens
per request, so the engine itself stays credential-agnostic and works
against any object store the caller can already authenticate against.
Icebird supports two access modes. If the caller already has a
metadata-file or snapshot URI, Icebird can read it directly from object
storage. If the caller needs current-table resolution, named branches,
or stronger coordination under concurrent writers, Icebird uses an
Iceberg catalog. Either way, the user and the agent each need only the
table reference and the credentials they already use for storage or
catalog access, with no shared backend on the hot path.

\hypertarget{agents-as-consumers}{%
\section{Agents as Consumers}\label{agents-as-consumers}}

Agents are the other major consumer of this trace data. Squirreling
loads directly into whatever JavaScript runtime the agent is already
executing in, whether that is Codex CLI's Node process, an ephemeral
\texttt{vm.Context} spawned per agent in a swarm, or a browser tab. No
separate query service sits in the path. For contrast, a Python query
engine behind an agent tool typically means a containerized service per
agent, with the per-turn cost of keeping it warm or paying its
cold-start. Squirreling instantiates in milliseconds, and JavaScript's
mature sandboxing makes per-agent isolation cheap. A production
deployment runs at hyperparam.app, where investigation turns issue 10 to
50 probes against the trace table (§5).

We test the agent-consumer case directly in a head-to-head with
DuckDB-WASM on the same Parquet corpus used in §3.2. An Anthropic
Haiku-class agent is handed an identical \texttt{run\_sql} tool backed
by each engine and asked to answer ten analyst-style questions about a
50,000-row agent-trace dataset, phrased the way a practitioner would ask
them, like ``which session had the roughest ride?'' or ``on turns where
the \texttt{web\_search} tool is invoked, how often does the turn end in
an error?'', rather than as direct SQL translations. Across 5
independent trials per task (50 runs per stack), both stacks resolve all
ten tasks correctly, but Squirreling does so at roughly a third of
DuckDB-WASM's mean cost per ten-task pass (Table 3).

\begin{table}[h]\centering
\begin{tabular}{@{}llll@{}}
\toprule
Stack & Correct & Median rounds & Cost per pass \\
\midrule
DuckDB-WASM & 50 / 50 & 2 & \$0.203 (\$0.10--\$0.28) \\
Squirreling & 50 / 50 & 2 & \$0.067 (\$0.05--\$0.08) \\
\bottomrule
\end{tabular}\end{table}

\emph{Table 3: Anthropic Haiku agent completing a ten-task analyst-style
suite over a 50,000-row agent-trace Parquet with} \texttt{run\_sql}
\emph{backed by each browser engine, 5 trials per task. Squirreling
0.12.16; DuckDB-WASM 1.33.1-dev45.0. Round cap R=10 with up to 3
additional inferences for the agent to submit an answer. Correctness is
scored against a reference answer pinned per task, with per-task numeric
tolerance and case-insensitive string matching. ``Cost per pass'' is
mean Anthropic Haiku list pricing summed across all tool-use rounds for
one ten-task pass, with min--max across the 5 passes in parentheses.}

The cost gap is almost entirely input tokens: across the 50 runs,
DuckDB-WASM consumed 3.67× the input tokens Squirreling did (876K vs
239K) against only 1.43× the output (28K vs 19K). The agent emits
roughly the same volume of SQL and prose on both stacks; what differs is
what the model is \emph{billed to read} on each subsequent call. The
Anthropic API bills \texttt{input\_tokens} for the full conversation
context on every \texttt{messages.create}, so each prior round's
\texttt{tool\_result} is re-billed on every later inference, and total
input grows quadratically in round count. Round count, in turn, is set
by the error surface. Squirreling's ``did you mean'' hints on near-miss
function names (e.g.~\texttt{lenght} \(\rightarrow\) \texttt{length})
and binder errors that enumerate available columns let the agent commit
to a working query sooner: on T7 (``\% turns that invoke a tool'')
DuckDB-WASM took a median 6 rounds and 42K mean input tokens vs
Squirreling's 1 round and 2.4K; on T4 (``\texttt{web\_search} turn error
rate''), 7 rounds and 72K vs 3 rounds and 7.1K. Steerability
\citep{liu2025agentic} at the error surface is the proximate cause, and
it compounds quadratically into the input-token bill.

Prompt caching narrows but does not close the gap: Anthropic
\texttt{cache\_control} breakpoints at tool-result boundaries take
prior-context tokens to 10\% of full price, dropping the R=5-vs-R=2
input-token ratio from 5× to 2.86×, but output tokens are still billed
at full rate and the round-count incentive remains.

\hypertarget{experience-and-limitations}{%
\section{Experience and Limitations}\label{experience-and-limitations}}

\textbf{Production deployment.} Each library ships with a live public
demo, reproducible against multi-gigabyte Parquet on Hugging Face or
Iceberg snapshots in S3. The stack is also used in production by
Hyperparam's own product at \url{https://hyperparam.app}, where
in-browser agents import Hyparquet and Squirreling into the user's
runtime; a typical investigation turn issues 10 to 50 probes against the
trace table, and subagent orchestration pushes that higher. Hyperparam
Desktop, an Electron app, adds persistent authentication for private S3
buckets and Iceberg catalogs behind corporate auth, resolving the
credential-refresh and CORS constraints for the private-data case
without giving up the in-place model. These deployments shaped the
per-cell laziness design, since agent turns demand tight LLM-cost
bounds, and the custom-fetch hook in Icebird, since credentials never
leave the caller's execution context.

\textbf{Where backend-free is the wrong choice.} Authentication against
private object stores is a real client-side concern, and the right
answer depends on the host runtime. A web tab needs short-lived
browser-vended tokens, an Electron app can persist credentials in the OS
keychain, which is the route Hyperparam Desktop takes for private S3,
and a Node CLI inherits whatever the shell already has. Aggregations
over tens of billions of rows, multi-table joins exceeding RAM, or
workloads that need server-side spill-to-disk are outside the operating
envelope; the terabyte join still belongs on a cluster. Per-cell
laziness bounds inference cost for \texttt{LIMIT}-bounded queries but
does not prevent an unbounded \texttt{SELECT\ llm(...)\ FROM\ traces};
rate limiting, cost preview, and per-UDF token budgets are open UX
problems.

\hypertarget{conclusion}{%
\section{Conclusion}\label{conclusion}}

AI applications run on the client now, in JavaScript, with agents and
humans sharing one process and reasoning over text data their owners do
not want to move. The data engineering stack, built for a server-side,
SQL-only, human-paced era, sits on the wrong side of the runtime
boundary for this class of application. We argue that data engineering
as a discipline needs to update for this new shape of data and the new
shape of queries that run on it.

\bibliographystyle{ACM-Reference-Format}
\bibliography{A_Query_Engine_for_the_Agents_v4}

@misc{deepview2026,
  author = {{The Deep View}},
  title = {{AI's compute crisis has reached a breaking point}},
  year = {2026},
  url = {https://www.thedeepview.com/articles/ai-s-compute-crisis-has-reached-a-breaking-point},
  note = {Accessed 2026}
}

@article{liu2025agentic,
  author = {Liu, Shu and Ponnapalli, Soujanya and Shankar, Shreya and Zeighami, Sepanta and Zhu, Alan and Agarwal, Shubham and Chen, Ruiqi and Suwito, Samion and Yuan, Shuo and Stoica, Ion and Zaharia, Matei and Cheung, Alvin and Crooks, Natacha and Gonzalez, Joseph E. and Parameswaran, Aditya G.},
  title = {Supporting Our {AI} Overlords: Redesigning Data Systems to be Agent-First},
  journal = {arXiv:2509.00997},
  year = {2025}
}

@article{tagliabue2025lakehouse,
  author = {Tagliabue, Jacopo and Bianchi, Federico and Greco, Ciro},
  title = {Trustworthy {AI} in the Agentic Lakehouse: from Concurrency to Governance},
  journal = {arXiv:2511.16402},
  year = {2025}
}

@article{kohn2022duckdbwasm,
  author = {Kohn, Andr{\'e} and Moritz, Dominik and Raasveldt, Mark and M{\"u}hleisen, Hannes and Neumann, Thomas},
  title = {{DuckDB-Wasm}: Fast Analytical Processing for the Web},
  journal = {Proceedings of the VLDB Endowment},
  volume = {15},
  number = {12},
  year = {2022}
}

@inproceedings{abadi2007materialization,
  author = {Abadi, Daniel J. and Myers, Daniel S. and DeWitt, David J. and Madden, Samuel R.},
  title = {Materialization Strategies in a Column-Oriented {DBMS}},
  booktitle = {ICDE},
  year = {2007}
}

@inproceedings{armbrust2021lakehouse,
  author = {Armbrust, Michael and Ghodsi, Ali and Xin, Reynold S. and Zaharia, Matei},
  title = {Lakehouse: A New Generation of Open Platforms That Unify Data Warehousing and Advanced Analytics},
  booktitle = {CIDR},
  year = {2021}
}

@inproceedings{raasveldt2019duckdb,
  author = {Raasveldt, Mark and M{\"u}hleisen, Hannes},
  title = {{DuckDB}: an Embeddable Analytical Database},
  booktitle = {SIGMOD},
  year = {2019}
}

@incollection{vohra2016parquet,
  author = {Vohra, Deepak},
  title = {Apache Parquet},
  booktitle = {Practical Hadoop Ecosystem},
  publisher = {Apress},
  year = {2016}
}

@misc{iceberg_spec,
  author = {{Apache Iceberg}},
  title = {{Apache Iceberg Table Spec v2}},
  year = {2024},
  url = {https://iceberg.apache.org/spec/}
}

@misc{apache_arrow,
  author = {{Apache Arrow}},
  title = {{Apache Arrow JavaScript Library (arrow-js)}},
  year = {2024},
  url = {https://arrow.apache.org/docs/js/}
}

@misc{moritz2020arquero,
  author = {Moritz, Dominik and Heer, Jeffrey},
  title = {{Arquero}: Query Processing and Transformation of Array-Backed Data Tables},
  howpublished = {Observable},
  year = {2020},
  url = {https://observablehq.com/@uwdata/arquero}
}

@inproceedings{shankar2024validators,
  author = {Shankar, Shreya and Zamfirescu-Pereira, J.D. and Hartmann, Bj{\"o}rn and Parameswaran, Aditya G. and Arawjo, Ian},
  title = {Who Validates the Validators? Aligning {LLM}-Assisted Evaluation of {LLM} Outputs with Human Preferences},
  booktitle = {UIST},
  year = {2024}
}

@misc{zaharia2024compound,
  author = {Zaharia, Matei and Khattab, Omar and Chen, Lingjiao and Davis, Jared Quincy and Miller, Heather and Potts, Chris and Zou, James and Carbin, Michael and Frankle, Jonathan and Rao, Naveen and Ghodsi, Ali},
  title = {The Shift from Models to Compound {AI} Systems},
  howpublished = {BAIR Blog},
  year = {2024},
  url = {https://bair.berkeley.edu/blog/2024/02/18/compound-ai-systems/}
}

@article{melnik2010dremel,
  author = {Melnik, Sergey and Gubarev, Andrey and Long, Jing Jing and Romer, Geoffrey and Shivakumar, Shiva and Tolton, Matt and Vassilakis, Theo},
  title = {{Dremel}: Interactive Analysis of Web-Scale Datasets},
  journal = {Proceedings of the VLDB Endowment},
  volume = {3},
  number = {1},
  year = {2010}
}

@article{he2021automatedlog,
  author = {He, Shilin and He, Pinjia and Chen, Zhuangbin and Yang, Tianyi and Su, Yuxin and Lyu, Michael R.},
  title = {A Survey on Automated Log Analysis for Reliability Engineering},
  journal = {ACM Computing Surveys},
  volume = {54},
  number = {6},
  articleno = {130},
  year = {2021},
  doi = {10.1145/3460345}
}

@inproceedings{alspaugh2014loganalysis,
  author = {Alspaugh, Sara and Chen, Beidi and Lin, Jessica and Ganapathi, Archana and Hearst, Marti A. and Katz, Randy},
  title = {Analyzing Log Analysis: An Empirical Study of User Log Mining},
  booktitle = {LISA},
  year = {2014},
  publisher = {USENIX Association}
}

@inproceedings{john2023datachat,
  author = {John, Rogers Jeffrey Leo and Bacon, Dylan and Chen, Junda and Ramesh, Ushmal and Li, Jiatong and Das, Deepan and Claus, Robert and Kendall, Amos and Patel, Jignesh M.},
  title = {{DataChat}: An Intuitive and Collaborative Data Analytics Platform},
  booktitle = {SIGMOD Companion},
  year = {2023},
  doi = {10.1145/3555041.3589678}
}

@article{dong2024agentops,
  author = {Dong, Liming and Lu, Qinghua and Zhu, Liming},
  title = {{AgentOps}: Enabling Observability of {LLM} Agents},
  journal = {arXiv:2411.05285},
  year = {2024}
}

@article{dai2024uqe,
  author = {Dai, Hanjun and Wang, Bethany Yixin and Wan, Xingchen and Dai, Bo and Yang, Sherry and Nova, Azade and Yin, Pengcheng and Phothilimthana, Phitchaya Mangpo and Sutton, Charles and Schuurmans, Dale},
  title = {{UQE}: A Query Engine for Unstructured Databases},
  journal = {arXiv:2407.09522},
  year = {2024}
}

@inproceedings{liu2025palimpzest,
  author = {Liu, Chunwei and Russo, Matthew and Cafarella, Michael and Cao, Lei and Chen, Peter Bailis and Chen, Zui and Franklin, Michael and Kraska, Tim and Madden, Samuel and Shahout, Rana and Vitagliano, Gerardo},
  title = {{Palimpzest}: Optimizing AI-Powered Analytics with Declarative Query Processing},
  booktitle = {CIDR},
  year = {2025}
}

@article{patel2025semanticoptimization,
  author = {Patel, Liana and Jha, Siddharth and Pan, Melissa and Gupta, Harshit and Asawa, Parth and Guestrin, Carlos and Zaharia, Matei},
  title = {Semantic Operators and Their Optimization: Enabling {LLM}-Based Data Processing with Accuracy Guarantees in {LOTUS}},
  journal = {Proceedings of the VLDB Endowment},
  year = {2025},
  doi = {10.14778/3749646.3749685}
}

\end{document}